\pdfoutput=1
\documentclass[11pt]{article}

\usepackage[]{ACL2023}

\usepackage{times}
\usepackage{latexsym}

\usepackage[T1]{fontenc}
\usepackage[utf8]{inputenc}

\usepackage{microtype}

\usepackage{inconsolata}

\usepackage{graphicx}
\usepackage{amsmath}
\usepackage{booktabs}
\usepackage{pifont}
\usepackage{makecell}
\usepackage{multirow}

\newcommand\mf[1]{\mathbf{#1}}

\newcommand\ttt[1]{\texttt{#1}}

\newcommand{\maskk}{\ttt{<mask>}}

\title{Improving Contrastive Learning of Sentence Embeddings from AI Feedback}

\author{Qinyuan Cheng, Xiaogui Yang, Tianxiang Sun, Linyang Li, Xipeng Qiu\footnotemark[2] \\
        School of Computer Science, Fudan University \\
        chengqy21@m.fudan.edu.cn
        }
\begin{document}
\maketitle
\begin{abstract}
Contrastive learning has become a popular approach in natural language processing, particularly for the learning of sentence embeddings.
However, the discrete nature of natural language makes it difficult to ensure the quality of positive and negative sample pairs generated through data augmentation methods.
Although supervised contrastive learning can produce more accurate sample pairs with human feedback labels, it still lacks fine-grained training signals.
In this paper, we propose to improve \textbf{C}ontrastive \textbf{L}earning of sentence embeddings from \textbf{AI} \textbf{F}eedback \textbf{(CLAIF)}.
Our method utilizes AI feedback from large pre-trained language models (LLMs) to construct sample pairs with fine-grained sample similarity scores to improve contrastive learning.
Besides, we combine human feedback and AI feedback to provide better supervision signals for supervised contrastive learning of sentence embeddings.
Experimental results show that our method achieves state-of-the-art performance on several semantic textual similarity (STS) and transfer learning tasks compared to other unsupervised and supervised contrastive learning methods.
\footnote{We will release our code and data at \url{https://github.com/xiami2019/CLAIF}}
\renewcommand{\thefootnote}{\fnsymbol{footnote}} %将脚注符号设置为fnsymbol类型，即特殊符号表示
\footnotetext[2]{Corresponding author.}

\end{abstract}

\section{Introduction}
Learning sentence embeddings with rich semantics is very important for many natural language processing tasks, such as semantic matching and information retrieval. 
Recently, pre-trained language models \cite{BERT,RoBERTa,PTM_Survey} provide a convenient way to get sentence embeddings. 
However, sentence embeddings directly generated by pre-trained language models show poor performance on semantic textual similarity (STS) tasks due to the representation degeneration problem \cite{DBLP:conf/iclr/GaoHTQWL19}. 
Therefore, finding ways to further improve pre-trained models to produce better sentence embeddings becomes an crucial and fundamental challenge in natural language processing.

% \citet{SentenceBERT} show that the regression objective function is effective for learning sentence embeddings. 
% 要解决的问题
Given the shortage of labeled data for sentence embedding learning, recent studies mainly focus on unsupervised methods, such as utilizing contrastive learning methods\cite{ConSERT, SimCSE, DiffCSE}.
Contrastive learning can be classified into two categories \cite{SCL}: supervised contrastive learning and unsupervised contrastive learning, depending on whether additional label information is utilized to construct positive and negative sample pairs.
However, the quality of positive and negative sample pairs in unsupervised contrastive learning can be difficult to ensure.
Recent studies also show that data augmentation strategies in unsupervised contrastive learning may introduce some bias like length information \cite{ESimCSE} and improper negatives \cite{Debiased_Negatives}.
While supervised contrastive learning methods can produce more accurate sample pairs by utilizing label information, such as using supervised datasets from natural language inference \cite{SimCSE}, it can only provide coarse-grained labels and lack fine-grained supervision signals.
We aruge that these limitations of current contrastive learning methods restrict further performance enhancement of sentence embeddings.

\begin{figure*}[!t]
\centering
\includegraphics[width=1.0\linewidth]{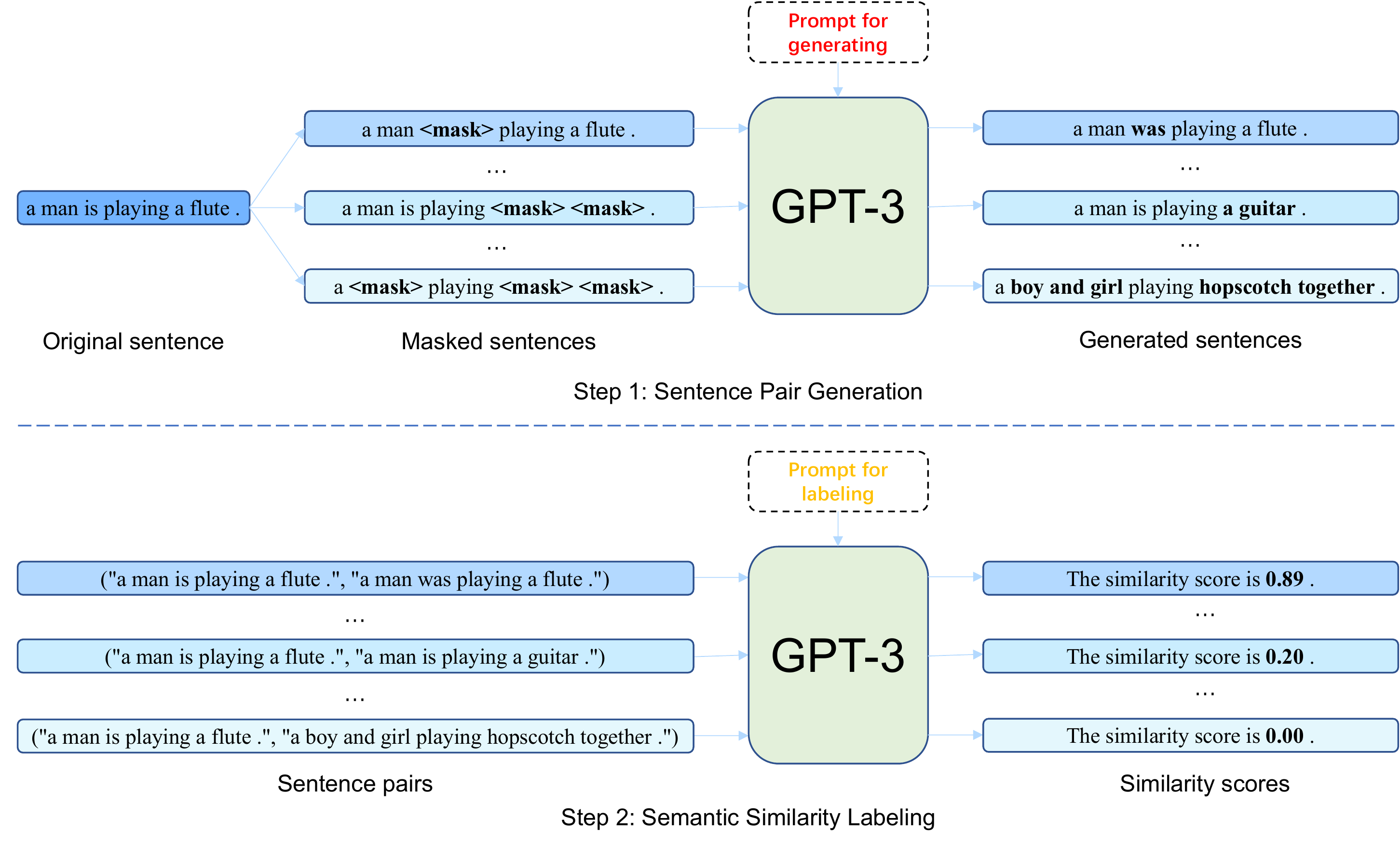}
\centering
\caption{Illustration of the sample pair generation process. The darker the color, the more information the sentence shares with the original sentence.}
\label{fig:generation_process}
\end{figure*}

% 受到的启发：From Dataset Generation
With the emergence of large pre-trained language models (LLMs) \cite{GPT3,ERNIE3.0,InstructGPT,OPT}, researchers hope powerful LLMs can help human train other AI models \cite{RLAIF}. 
One way is to use LLMs to generate datasets using for zero-shot learning \cite{Dino, ZeroGen, SuperGen}. 
These methods all use predefined labels and task descriptions to generate training inputs, instead of utilizing AI feedback as supervision signals.
Therefore, these method are not suitable for tasks whose labels are continuous values and may lead to lack of diversity in training samples.
Inspired by these studies, we hope to exploit the capability of LLMs to address shortcomings in contrastive learning of sentence embeddings.
% \citet{Dino} generate a STS dataset using three discrete labels: \emph{1: mean the same things, 0.5: somewhat similar, 0: on completely different topics}. 
% However, these discrete labels can not reflect fine-grained semantic similarities between sentences and may not fully leverage the linguistic knowledge of LLMs.

We propose to improve \textbf{C}ontrastive \textbf{L}earning of sentence embeddings from \textbf{AI} \textbf{F}eedback \textbf{(CLAIF)}.
Specifically, we design a two-step sample pair generation method to produce high quality sentence pairs and fine-grained semantic similarity scores using AI feedback from GPT-3, as shown in Figure \ref{fig:generation_process}.
In the first step, we mask some words in a sentence with different mask rates and then use GPT-3 to generate new sentences based on the remaining information in the masked sentence. 
Then we combine the generated sentences and the original sentence to construct sentence pairs.
In this way, we can use the mask rate to control the amount of sharing information between two sentences in a pair, which will produce sentence pairs with different semantic similarities.
In the second step, we utilize GPT-3 to generate semantic similarity scores for sentence pairs.
\textbf{These scores are the AI feedback on sample similarity.}
Since the semantic change caused by reconstructing a masked sentence is difficult to measure, we leverage the linguistic knowledge of LLMs to generate the semantic similarity score.
The diversity of AI feedback similarity scores ensured by the sentence pair generation process in the first step.
At last we use our generated sample pairs and similarity scores to train the model for sentence embeddings.

In addition to using AI feedback alone, we also combine human feedback and AI feedback by introducing AI feedback into supervised contrastive learning of sentence embeddings which needs human feedback labels to generate positive sample pairs.
We use the AI feedback similarity score for the positive sample pair as a soft label to replace the one-hot label in InfoNCE loss \cite{MOCO}.
We term our loss Soft InfoNCE.
This process can be referred to as contrastive learning of sentence embeddings from human and AI feedback (CLHAIF).

We conduct extensive experiments to show the effectiveness of our method. 
Sentence embeddings learned with CLAIF and CLHAIF achieve state-of-the-art performance on standard semantic textual similarity tasks and outperform strong baselines on transfer learning tasks.
We also find that CLAIF results in significant improvements to the cross-encoder architecture for the sentence-pair modeling task.

\begin{table*}[t]
\centering
\small
\setlength{\tabcolsep}{8pt}
\begin{tabular}{lcccc}
\toprule
\textbf{Feedback Source} & 
\textbf{Positive Pair} &
\textbf{Negative Pair} &
% \textbf{Label Type} &
\textbf{Loss Function}
\\
\midrule
\makecell[l]{Zero Feedback \\ (CLZF)} & 
$(x_i, x_i^{\prime})$ &
% $(x_i, x_i^{\prime})$ &
$\left\{(x_i, x_j)\mid{x_j}\in{X}, i \neq j \right\}$ &
% Hard & 
\makecell[{}{p{4cm}}]{
InfoNCE \cite{CPC, MOCO, SimCSE}, \\
NT-Xent \cite{SimCLR}
}  \\
\midrule
\makecell[l]{Human Feedback \\ (CLHF)} & 
$(x_i, x_i^+)$ & 
$\left\{(x_i, x_i^-),(x_i, x_j)\mid{x_j}\in{X}, i \neq j \right\}$ &
% $(x_i, x_i^-),(x_i, x_j)$ &
% Hard & 
\makecell[{}{p{4cm}}]{
SupCon \cite{SCL}, \\
InfoNCE \cite{SimCSE}, \\
KNN-Contrastive \cite{zhou-etal-2022-knn}
}\\
\midrule
\makecell[l]{AI Feedback \\ (CLAIF)} & 
$(x_i, x_i^{\prime}, y_i)$ & 
$(x_i, x_i^{\prime}, y_i)^*$ & 
% Soft & 
\makecell[l]{Mean Squared Error} \\
\midrule
\makecell[l]{Human and AI Feedback \\ (CLHAIF)} & 
$(x_i, x_i^+, y_i)$ & 
$\left\{(x_i, x_i^-),(x_i, x_j)\mid{x_j}\in{X}, i \neq j \right\}$ &
% $(x_i, x_i^-),(x_i, x_j)$ &
% Soft & 
\makecell[l]{Soft InfoNCE}   \\
\bottomrule
\end{tabular}
\caption{
The details of contrastive learning from different feedback.
$X$ is the full set containing all samples and $x_i$ is the i-th sample of $X$, such as a sentence or an image.
$x_i^{\prime}$ is an augmented sample obtained by using some data augmentation strategies to $x_i$.
$x_i^+$ and $x_i^-$ are postive sample and negative sample of $x_i$ picked by human feedback information, such as class label information.
$y_i$ is the AI feedback sample similarity score for the i-th sample pair.
$*$: CLAIF does not explicitly construct positive and negative pairs, sample pairs with high simiarity scores can be seen as positive pairs and those with low scores can be seen as negative pairs. 
} 
\label{tab:cl_from_different_fb}
\end{table*}

Our main contributions are as follows:
\begin{itemize}
    \item We propose to improve contrastive learning of sentence embeddings from AI feedback (CLAIF) and achieve state-of-the-art performance on several semantic textual similarity tasks and transfer learning tasks. 
    % To the best of our knowledge, we are the first to introduce AI feedback into contrastive learning. 
    \item We construct a semantic textual similarity dataset with high quality sentence pairs and fine-grained AI feedback similarity scores using large pre-trained language models.
    \item We propose a method to incorporate human feedback and AI feedback to provide better supervision for contrastive learning of sentence embeddings.
    \item Experimental results show the scalability of CLAIF, which is cheaper and more efficient than collecting data from human feedback.
    
    % \item We propose CLAIF which improves both unsupervised and supervised contrastive learning of sentence embeddings from AI feedback and achieves state-of-the-art performance on several semantic textual similarity tasks and transfer tasks.
    % \item We construct a semantic textual similarity dataset from scratch only rely on AI feedback, which has high quality sentence pairs and fine-grained similarity labels.
    % \item We add AI feedback labels to natural language inference (NLI) datasets which are commonly used in supervised contrastive learning of sentence embeddings.
    % \item We will release our code and generated datasets to encourage the future work of sentence embeddings.
\end{itemize}

\section{Understanding Contrastive Learning from Different Feedback}
In this section, we categorize contrastive learning methods into four categories according to their feedback sources.
We summarize the details of contrastive learning from different feedback in Table \ref{tab:cl_from_different_fb}, including their feedback types, sample pairs construction methods and representative loss functions.

\subsection{Contrastive Learning from Zero Feedback}

Traditional contrastive learning is used for self-supervised representation learning \cite{Contrastive_loss, MOCO}.
These methods construct positive and negative sample pairs using data augmentation strategies without any human feedback.
For example, in natural language processing, \citet{SimCSE} construct positive sample pairs by doing the dropout operation twice for the same sentence and negative pairs by combining with another sentences.
We refer to these methods as Contrastive Learning from Zero Feedback (CLZF).
The most common loss function for CLZF is InfoNCE \cite{CPC}.
\citet{SimCLR} propose NT-Xent loss, which can be seen as a variant of InfoNCE.
However, due to the discrete nature of natural language, it is hard to find effective and unbiased data augmentation strategies to construct high quality sample pairs.

\subsection{Contrastive Learning from Human Feedback}
\label{section:CLHF}
Recently, \citet{SCL} propose to use label information to construct positive sample pairs.
In sentence embeddings, \citet{SimCSE} use premise-hypothesis pairs with entailment relationship from natural language inference (NLI) datasets as positive sample pairs and still use InfoNCE for training.
Since these methods leverage label information from human, we refer to them as Contrastive Learning from Human Feedback (CLHF).
With the help of label information, some new losses can be used in CLHF, like SupCon \cite{SCL} and KNN-Contrastive \cite{zhou-etal-2022-knn}.
Although CLHF can construct more accurate sample pairs, it still lacks fine-grained supervision signals.
For example, in InfoNCE, all positive pairs have a label of 1.
But there are also differences in the similarity between different positive sample pairs.

\subsection{Contrastive Learning from AI Feedback}
Measuring the similarity of sample pairs in contrastive learning is a laborious task.
However, thanks to emergence of LLMs, we can use LLMs to measure the similarity of sample pairs and use the AI feedback as our training signals.
We refer to this approach as Contrastive Learning from AI Feedback (CLAIF).
CLAIF does not need to explicitly construct positive and negative sample pairs because each sample pair has a fine-grained label.
We use mean squared error (MSE) loss for the training of CLAIF in this work.
% We summarize the characteristics of different types of contrastive learning in Table \ref{tab:cl_from_different_fb}.
% In this paper we get all AI feedback from OpenAI's GPT-3.

\subsection{Contrastive Learning from Human and AI Feedback}
Besides contrastive learning from AI feedback, we propose to combine human and AI feedback to produce better supervision signals when they are both available.
We call this category contrastive learning from human and AI feedback (CLHAIF) and we propose a soft InfoNCE loss for the training of CLHAIF.
We hope to use fine-grained AI feedback to refine the coarse-grained signals in current CLHF methods.

\section{Methodology}
\label{sec:method}
In this section, we first introduce our method to generate sample pairs and the training process of CLAIF.
In order to obtain high quality sentence pairs with diverse and fine-grained similarity scores, we propose a two-step sample pair generation method: \textbf{Sentence Pair Generation} and \textbf{Semantic Similarity Labeling}.
The generation process is shown in Figure \ref{fig:generation_process}.
We use these sample pairs to train language models like BERT and RoBERTa.
Then we introduce CLHAIF, which combines human and AI feedback in contrastive learning of sentences embeddings.

\subsection{Sentence Pair Generation}
\label{sec:sentence_pair_generation}
We use unpaired sentences from the training set of STS Benchmark \cite{STSb} as our original sentences.
As shown in Figure \ref{fig:generation_process}, we first mask some words of the original sentence \emph{"a man is playing a flute."} with different mask rates using the \maskk \ token, in order to delete some information in the original sentence.
The more words that are masked, the less information is left.
We use the depth of color to indicate the degree of information sharing between two sentences in Figure \ref{fig:generation_process}.
Then we write a task description prompt to steer GPT-3 to generate new sentences based on masked sentences.
We provide our task descriptions in Appendix \ref{sec:task_descriptions}. 
% \emph{"Replace all <mask> tokens in "<masked sentence>" to make a new sentence. The new sentence is:"}.
% The prompt for unmasked sentences (mask rate is 0) is: \emph{"Write two sentences that mean the same thing. Sentence 1: "<original sentence>" Sentence 2:"}.
To increase the diversity of generated sentences, we merge adjacent \maskk \ tokens in 50\% of masked sentences into one \maskk \  token.
Then we combine the original sentence with each generated sentence to construct sentence pairs.
% This process is formalized as follows:
% \begin{gather}
%     x \xrightarrow{random\ mask} x_{masked} \nonumber \\
%     GPT(prompt_{step1},x_{masked})\xrightarrow{} x^{\prime} \nonumber
% \end{gather}
% where $x$ is the original sentence, $x_{masked}$ is the masked sentence and $x^{\prime}$ is the new generated sentence.
% The generated sentence pair is $(x, x^{\prime})$.

\subsection{Semantic Similarity Labeling}
\label{sec:semantic_similarity_labeling}
\begin{figure}[t]
\centering
\includegraphics[width=1.0\linewidth]{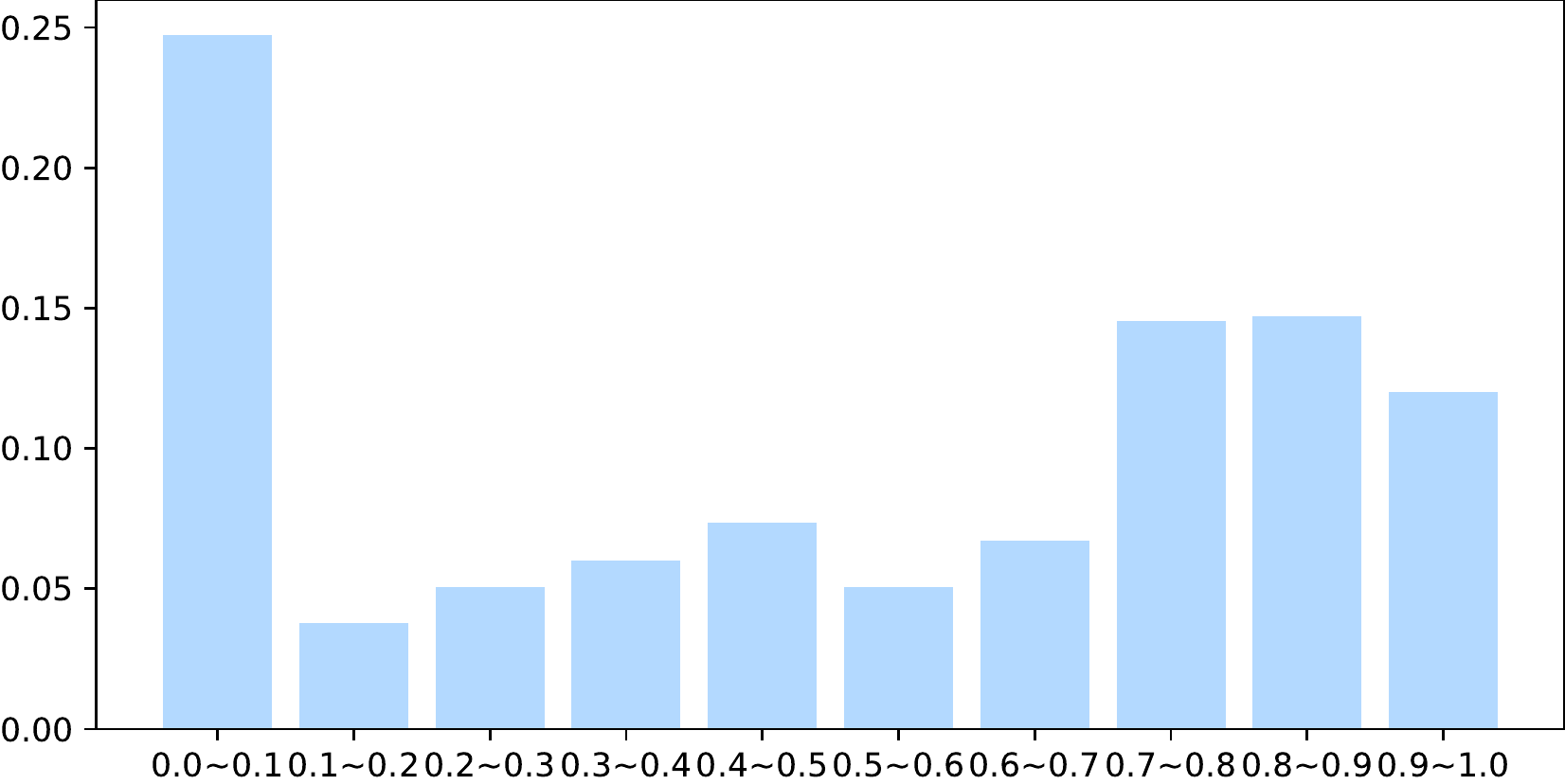}
\centering
\caption{The score distribution of our generated sample pairs. The x-axis is the similarity score and the y-axis is the percentage of the score.}
\label{fig:data_distribution}
\end{figure}

In this step, we label the semantic similarity score for each sentence pair using AI feedback from GPT-3.
The similarity score ranges from 0 to 1, where a score of 1 means that the semantic of the two sentences are exactly the same, and a score of 0 means that the semantic of the two sentences are completely different.
We write a task description prompt to steer GPT-3 to generate a similarity score between 0 and 1 for each sample pair generated in step 1.
% \emph{"The similarity score for two sentences is in the range from 0.0 to 1.0, 0.0 means completely different and 1.0 means almost the same. Now given two sentences '<sentence1>' and  '<sentence2>', please give a similarity score for these two sentences: The similarity score for these two sentences is"}
% This process is formalized as follows:
% \begin{gather}
%     GPT(prompt_{step2}, x, x^{\prime})\xrightarrow{} y \nonumber
% \end{gather}
% where $x$ and $x^{\prime}$ are from the sentence pair $(x,\ x^{\prime})$ and $y$ is the generated similarity score.
The first step ensures the diversity of semantic similarity scores.
As illustrated in Figure \ref{fig:data_distribution}, the generated scores are diverse and distributed in the value range from 0 to 1.

\subsection{Training on Generated Pairs}
\label{section:Further-Pre-training}
With the generated sample pairs, we train a language model as the sentence encoder to get better sentence embeddings.
Given diverse sentence pairs which have fine-grained similarity scores, we do not need to explicitly construct positive and negative sample pairs.
Therefore, we directly use the mean squared error (MSE) loss to fit the cosine similarity of each sentence pair to its AI feedback similarity score:
\begin{gather}
    \mathcal{L}=\frac{1}{N}\sum_{i=1}^{N}\big[\cos\left(\mf{h}_{i}, \mf{h}_{i}^{\prime}\right)-y_{i}\big]^{2}
\end{gather}
where $N$ is the batch size,
$\mf{h}_{i}$ and $\mf{h}_{i}^{\prime}$ are two sentence embeddings of the i-th sentence pair $(x_{i}, x_{i}^{\prime})$ encoded by the model, $y_i$ is the corresponding similarity score and 
$\cos$ means the calculation of cosine similarity.
During inference, we use the cosine similarity of two sentence embeddings as their semantic similarity score.

% \subsection{Feedback from Cross-Encoder}
% Sentence encoders we introduced in section \ref{section:Further-Pre-training} are bi-encoders. 
% Cross-encoder is another common architecture for sentence-pair modelling.
% It directly encodes the sequence concatenated by two sentences and then predicts a similarity score.
% Previous studies \cite{Augmented_SBERT, TransEncoder, ERNIE-Search} show that cross-encoders usually outperform bi-encoders.
% We find that our generated dataset can significantly improve the performance of cross-encoders on STS tasks, with the help of fine-grained labels.
% Therefore, we use the feedback from cross-encoders to further improve the performance of sentence embeddings learned with unsupervised CLAIF.

% We use the binary cross-entropy (BCE) loss to train cross-encoders with the same backbone models as our bi-encoders:
% \begin{gather}
%     \mathcal{L}=-\frac{1}{N}\sum_{i=1}^{N}l_i \\
%     l_i=y_{i}\log{\sigma(\hat{y_i})}+(1-y_{i})\log{(1-\sigma(\hat{y_i}))} \nonumber
% \end{gather}
% where $N$ is the batch size,
% $\hat{y_i}$ is the predicted score of the i-th sentence pair,
% $y_i$ is the corresponding similarity score and $\sigma$ is the sigmoid function.

% Then we use the trained cross-encoder to generate pseudo labels for all sentence pairs and use these pseudo labels and the MSE loss to train a bi-encoder.
% The bi-encoder here is initialized from the model trained in Section \ref{section:Further-Pre-training}.

\subsection{Combining Human Feedback and AI Feedback}
In this section, we mainly study the cooperation of human and AI models to provide better training signals for contrastive learning, which we called CLHAIF.
\citet{SentenceBERT} use supervised NLI datasets to learn sentence embeddings.
\citet{SimCSE} construct positive and hard negative sample pairs for contrastive learning leveraging label information of NLI datasets, achieving significant improvements.
However, as we mentioned in Section \ref{section:CLHF}, CLHF does not distinguish between different positive sample pairs and assigns label of 1 for all positive pairs.
In this way, all positive sample pairs are pulled together with the same extent in contrastive learning, ignoring differences in similarity between different positive pairs.
Therefore, we use AI feedback to refine these coarse-grained supervision signals.

At first, we use the semantic similarity labeling step in Section \ref{sec:semantic_similarity_labeling} to generate AI feedback similarity scores for sentence pairs constructed from supervised NLI datasets: SNLI \cite{SNLI} and MNLI \cite{MNLI}.
Following \citet{SimCSE}, we construct sample pairs using the label information.
For the i-th sample of the NLI dataset, we can obtain two sentence pairs $(x_i,x_i^+)$ and $(x_i,x_i^-)$, where $x_i$ is the premise, $x_i^+$ and $x_i^-$ are entailment and contradiction hypothesis.
$(x_i,x_i^+)$ is the positive pair and $(x_i,x_i^-)$ is the hard negative pair.

In order to incorporate AI feedback, we propose soft InfoNCE loss by replacing the one-hot label with the AI feedback score as the soft label:
\begin{gather}
    \mathcal{L}=-\frac{1}{N}\sum_{i=1}^{N}l_i \\
    l_i=y_i \log \frac{e^{\cos(\mf{h}_i,\mf{h}_i^+ )/ \tau }}{\sum_{j=1}^N\left(e^{\cos(\mf{h}_i,\mf{h}_j^+)/\tau}+e^{\cos(\mf{h}_i,\mf{h}_j^-)/ \tau}\right)} \nonumber
\end{gather}
where N is the batch size, 
$\mf{h}_i$, $\mf{h}_i^+$ and $\mf{h}_i^-$ are sentence embeddings of $x_i$, $x_i^+$ and $x_i^-$,
$y_i$ is the AI feedback similarity score for the positive pair $(x_i,x_i^+)$ and $\tau$ is the temperature parameter.

\section{Experiments}

\subsection{Evaluation Datasets}
We conduct extensive experiments on seven semantic textual similarity (STS) tasks and seven transfer learning tasks.
The STS tasks include STS 2012-2016 \cite{STS2012,STS2013,STS2014,STS2015,STS2016}, STS Benchmark \cite{STSb} and SICK-Relatedness \cite{SICK-R}.
The transfer learning tasks include MR \cite{MR}, CR \cite{CR}, SUBJ \cite{SUBJ}, MPQA \cite{MPQA}, SST-2 \cite{SST}, TREC \cite{TREC} and MRPC \cite{MRPC}.

Following \citet{SimCSE}, for STS tasks, we calculate the Spearman's correlation between the cosine similarity of sentence embeddings and the golden similarity scores from STS datasets.
For transfer learning tasks, we train a logistic regression classifier based on fixed sentence embeddings and follow the default settings of SentEval \cite{SentEval}.
% \footnote{\url{https://github.com/facebookresearch/SentEval}} 公开版再加链接
We use the same evaluation script as \citet{SimCSE} to calculate metrics.

\subsection{Baselines}
We compare our method with some strong baselines among three types of sentence embedding methods: 

\noindent \textbf{Post-processing methods}: 
These methods adopt some post-processing operations to enhance sentence embeddings which do not need to further train the backbone model.
We use BERT-whitening \cite{BERT-Whitening}, BERT-flow \cite{BERT-Flow} and prompt based BERT \cite{PromptBERT} as baselines.

\noindent \textbf{Training methods}:
These methods use additional data to further train the backbone model for better sentence embeddings.
We use SBERT \cite{SentenceBERT}, ConSERT \cite{ConSERT}, SimCSE \cite{SimCSE}, DiffCSE \cite{DiffCSE} and PromptBERT \cite{PromptBERT} as baselines.

\noindent \textbf{Dataset-generation based methods}: 
Some studies generate datasets from LLMs for sentence embedding learning.
We use Dino \cite{Dino} as our baseline.
Dino generates sentence pairs based on three discrete similarity labels using GPT2-XL.
For a fair comparison, we re-implement Dino using GPT-3 in our experiments.

\subsection{Implementation Details}
\label{sec:implementation_details}
\textbf{Choice of large pre-trained language models}: In our experiments, we get all AI feedback from text-davinci-003, which is the latest version of GPT-3.
We access text-davinci-003 through the OpenAI API.

\noindent \textbf{Sample pair generation}: 
We use nine mask rates for each original sentence in sentence pair generation: \emph{0.0, 0.1, 0.2, 0.3, 0.4, 0.5, 0.6, 0.7, 0.8}.
% we use a label smoothing technique for generated similarity scores \cite{Dino}.
For CLAIF, we use unpaired sentences from the training set of STS-B as original sentences to construct sentence pairs from scratch and randomly sample two other sentences for each original sentence to construct two sentence pairs with a similarity score of 0.
For CLHAIF, following previous studies \cite{SimCSE, PromptBERT}, we use the SNLI and MNLI datasets to construct sentence pairs and add a AI feedback similarity score for each sentence pair.
We only use the AI feedback scores for positive pairs in our experiments of CLHAIF.
Besides, to demonstrate the scalability of CLAIF, we use sentence pairs constructed from STS-B and from NLI datasets for the training of CLAIF, which we called CLAIF$_{\textrm{scaled}}$.
We list statistics of some datasets used for different methods in Table \ref{tab:statistic}.

\begin{table}[t]
\centering
\small
\resizebox{1.0\columnwidth}{!}{
\begin{tabular}{lcc}
\hline
\textbf{Dataset} & \textbf{Sample Number} & \textbf{Sample Type}\\
\hline
Wiki-1M & 1,000,000 & sentence \\
NLI & 275,601 & sentence triplet \\
Dino & 83,497 & sentence pair \\
CLAIF & 113,773 & sentence pair  \\
CLAIF$_{\textrm{scaled}}$ & 1,215,618 & sentence pair \\
\hline
\end{tabular}
}
\caption{Statistics of datasets for different settigns. Wiki-1M is used by CLZF methods. NLI is used by CLHF methods. We use CLAIF and CLAIF$_{\textrm{scaled}}$ to refer to our generated datasets here.}
\label{tab:statistic}
\end{table}

\noindent \textbf{Training details}:
We use the base version of the pre-trained language model BERT \cite{BERT} and RoBERTa \cite{RoBERTa} as our backbone models.
We use the development set of STS-B as our validation set.
In CLAIF, we use the mean pooling strategy to get sentence embeddings for BERT and RoBERTa.
For CLHAIF, we take the same pooling strategy as the corresponding baseline.
Other implementation details are in Appendix \ref{sec:training_details}.

\begin{table}[t]
\centering
% \small
% \setlength{\tabcolsep}{4pt}
% \resizebox{0.8\columnwidth}{!}{
\begin{tabular}{lc}
\hline
\textbf{Model} & \textbf{SentEval Avg.}\\
\hline
SimCSE$_{\textrm{BERT}}$ & 85.81 \\
PromptBERT & 85.49 \\
DiffCSE$_{\textrm{BERT}}$ & \textbf{86.86} \\
CLAIF$_{\textrm{BERT}}$ & 86.62 \\
\hline
SimCSE$_{\textrm{RoBERTa}}$ & 84.84 \\
PromptRoBERTa & 87.36 \\
DiffCSE$_{\textrm{RoBERTa}}$ & 87.04 \\
CLAIF$_{\textrm{RoBERTa}}$ & \textbf{87.99} \\
\hline
SimCSE$_{\textrm{BERT-supervised}}$ & 86.51 \\
\hspace{1em} w/ CLHAIF & 86.73 \\
PromptBERT$_{\textrm{supervised}}$ & 86.98 \\
\hspace{1em} w/ CLHAIF & \textbf{87.09} \\
\hline
SimCSE$_{\textrm{RoBERTa-supervised}}$ & 88.08 \\
\hspace{1em} w/ CLHAIF & 88.82 \\
PromptRoBERTa$_{\textrm{supervised}}$ & 89.11 \\
\hspace{1em} w/ CLHAIF & \textbf{89.27} \\
\hline
CLAIF$_{\textrm{scaled-BERT}}$ & 87.15  \\
CLAIF$_{\textrm{scaled-RoBERTa}}$ & \textbf{89.44} \\
\hline
\end{tabular}
% }
\caption{The performance comparison of CLAIF and CLHAIF on transfer learning tasks. SentEval Avg is the average accuracy on seven transfer learning datasets from SentEval.}
\label{tab:result_transfer_avg}
\end{table}

\begin{table*}[!ht]
\centering
\renewcommand\arraystretch{1.1}
\small
\setlength{\tabcolsep}{9.5pt}

\begin{tabular}{lcccccccc}
\hline
\textbf{Model} & \textbf{STS12} & \textbf{STS13} & \textbf{STS14} & \textbf{STS15} & \textbf{STS16} & \textbf{STS-B} & \textbf{SICK-R} & \textbf{Avg.}\\
\hline
\hline
\multicolumn{9}{c}{\textit{BERT-base}} \\
\hline
BERT-flow\(^\dagger\) & 58.40 & 67.10 & 60.85 & 75.16 & 71.22 & 68.66 & 64.47 & 66.55 \\
BERT-whitening\(^\dagger\)  & 57.83 & 66.90 & 60.90 & 75.08 & 71.31 & 68.24 & 63.73 & 66.28 \\
Prompt based BERT\(^\dagger\) & 60.96 & 73.83 & 62.18 & 71.54 & 68.68 & 70.60 & 67.16 & 67.85 \\
ConSERT\(^\dagger\) & 64.64 & 78.49 & 69.07 & 79.72 & 75.95 & 73.97 & 67.31 & 72.74 \\
SimCSE\(^\dagger\) & 68.40 & 82.41 & 74.38 & 80.91 & 78.56 & 76.85 & 72.23 & 76.25 \\
DiffCSE\(^\ddagger\) & 72.28 & 84.43 & 76.47 & 83.90 & 80.54 & 80.59 & 71.23 & 78.49 \\
PromptBERT\(^\dagger\) & 71.56 & \textbf{84.58} & \textbf{76.98} & 84.47 & 80.60 & 81.60 & 69.87 & 78.54 \\
\hline 
Dino$_{\textrm{GPT-3}}$ & \textbf{72.61} & 81.92 & 75.09 & 80.42 & 76.26 & 77.10 & 70.43 & 76.26 \\
\hline
CLAIF & 70.62 & 81.51 & 76.29 & \textbf{85.05} & \textbf{81.36} & \textbf{84.34} & \textbf{78.22} & \textbf{79.63} \\
\hline
\hline

\multicolumn{9}{c}{\textit{RoBERTa-base}} \\
\hline
RoBERTa-whitening\(^\dagger\) & 46.99 & 63.24 & 57.23 & 71.36 & 68.99 & 61.36 & 62.91 & 61.73 \\
SimCSE\(^\dagger\) & 70.16 & 81.77 & 73.24 & 81.36 & 80.65 & 80.22 & 68.56 & 76.57 \\
DiffCSE\(^\ddagger\) & 70.05 & 83.43 & 75.49 & 82.81 & 82.12 & 82.38 & 71.19 & 78.21 \\
PromptRoBERTa\(^\dagger\) & \textbf{73.94} & \textbf{84.74} & \textbf{77.28} & 84.99 & 81.74 & 81.88 & 69.50 & 79.15 \\
\hline
Dino$^\S$ & 70.27 & 81.26 & 71.25 & 80.49 & 77.18 & 77.82 & 68.09 & 75.20 \\
Dino$_{\textrm{GPT-3}}$ & 71.24 & 81.55 & 75.67 & 81.42 & 78.77 & 80.10 & 71.31 &77.15 \\
\hline
CLAIF & 68.33 & 82.26 & 77.00 & \textbf{85.18} & \textbf{83.43} & \textbf{85.05} & \textbf{78.02} & \textbf{79.90} \\
\hline
\end{tabular}
\caption{The performance comparison of CLAIF on STS tasks.
\(\dagger\): results from \cite{PromptBERT}. \(\ddagger \): results from \cite{DiffCSE}. $\S $: results from \cite{Dino}. Other results are from our experiments.
We bold the highest results among models with the same backbone.
} \label{tab:unsup_claif}
\end{table*}

\begin{table*}[!ht]
\centering
\renewcommand\arraystretch{1.1}
\small
\setlength{\tabcolsep}{3.8pt}

\begin{tabular}{lcccccccc}
\hline
\textbf{Model} & \textbf{STS12} & \textbf{STS13} & \textbf{STS14} & \textbf{STS15} & \textbf{STS16} & \textbf{STS-B} & \textbf{SICK-R} & \textbf{Avg.}\\
\hline
\hline
\multicolumn{9}{c}{\textit{BERT-base}} \\
\hline
% InferSent-GloVe \(^\dagger\) & 52.86 & 66.75 & 62.15 & 72.77 & 66.87 & 68.03 & 65.65 & 65.01 \\
SBERT\(^\dagger\)  & 70.97 & 76.53 & 73.19 & 79.09 & 74.30 & 77.03 & 72.91 & 74.89 \\
SBERT-flow\(^\dagger\)  & 69.78 & 77.27 & 74.35 & 82.01 & 77.46 & 79.12 & 76.21 & 76.60 \\
SBERT-whitening\(^\dagger\)  & 69.65 & 77.57 & 74.66 & 82.27 & 78.39 & 79.52 & 76.91 & 77.00 \\
ConSERT\(^\dagger\)  & 74.07 & 83.93 & 77.05 & 83.66 & 78.76 & 81.36 & 76.77 & 79.37 \\
SimCSE\(^\dagger\)  & \textbf{75.30} & 84.67 & 80.19 & 85.40 & 80.82 & 84.25 & 80.39 & 81.57 \\
\hspace{1em} w/ CLHAIF & 74.86$_{\downarrow 0.44}$ & 85.09$_{\uparrow 0.42}$ & 81.24$_{\uparrow 1.05}$ & 85.96$_{\uparrow 0.56}$ & 81.33$_{\uparrow 0.51}$ & 84.96$_{\uparrow 0.71}$ & \textbf{81.36}$_{\uparrow 0.97}$ & 82.08$_{\uparrow 0.51}$ \\
PromptBERT$^*$  & 75.10 & 85.54 & 80.58 & 86.00 & 81.24 & 84.57 & 80.36 & 81.91 \\
\hspace{1em} w/ CLHAIF & 75.03$_{\downarrow 0.07}$ & \textbf{85.88}$_{\uparrow 0.34}$ & \textbf{81.48}$_{\uparrow 0.90}$ & 86.33$_{\uparrow 0.33}$ & 81.40$_{\uparrow 0.16}$ & 84.93$_{\uparrow 0.36}$ & 80.98$_{\uparrow 0.62}$ & 82.29$_{\uparrow 0.38}$ \\
\hline 
CLAIF$_{\textrm{scaled}}$ & 74.36 & 85.07 & 80.64 & \textbf{87.21} & \textbf{83.36} & \textbf{86.26} & 79.68 & \textbf{82.37} \\
\hline
\hline

\multicolumn{9}{c}{\textit{RoBERTa-base}} \\
\hline
SRoBERTa\(^\dagger\)  & 71.54 & 72.49 & 70.80 & 78.74 & 73.69 & 77.77 & 74.46 & 74.21 \\
SRoBERTa-whitening\(^\dagger\)  & 70.46 & 77.07 & 74.46 & 81.64 & 76.43 & 79.49 & 76.65 & 76.60 \\
SimCSE\(^\dagger\)  & \textbf{76.53} & 85.21 & 80.95 & 86.03 & 82.57 & 85.83 & 80.50 & 82.52 \\
\hspace{1em} w/ CLHAIF & 76.23$_{\downarrow 0.30}$ & 85.46$_{\uparrow 0.25}$ & 81.48$_{\uparrow 0.53}$ & 86.47$_{\uparrow 0.44}$ & 83.40$_{\uparrow 0.83}$ & 85.93$_{\uparrow 0.10}$ & \textbf{80.95}$_{\uparrow 0.45}$ & 82.85$_{\uparrow 0.33}$ \\
PromptRoBERTa$^*$ & 76.41 & 85.64 & 82.11 & 86.18 & 82.71 & 85.74 & 79.95 & 82.68 \\
\hspace{1em} w/ CLHAIF & 76.26$_{\downarrow 0.15}$ & \textbf{86.01}$_{\uparrow 0.37}$ & \textbf{82.83}$_{\uparrow 0.72}$ & 86.70$_{\uparrow 0.52}$ & 82.94$_{\uparrow 0.23}$ & \textbf{86.04}$_{\uparrow 0.30}$ & 80.55$_{\uparrow 0.60}$ & \textbf{83.05}$_{\uparrow 0.37}$ \\
\hline
CLAIF$_{\textrm{scaled}}$ & 72.58 & 84.50 & 79.48 & \textbf{86.92} & \textbf{84.19} & 85.85 & 79.64 & 81.88 \\
\hline
\end{tabular}
\caption{The performance comparison of CLHAIF on STS tasks.
\(\dagger\): results from \citet{PromptBERT}. 
Other results are from our experiments.
$*$: The results of PromptBERT and PromptRoBERTa are obtained by running official code of \citet{PromptBERT} with recommended hyperparameters.
} \label{tab:sup_claif}
\end{table*}

\subsection{Main Results}
\noindent \textbf{Semantic Textual Similarity} \ 
We compare CLAIF with methods which do not use additional labeled datasets for training, including CLZF methods and dataset generation methods.
The results of CLAIF on STS tasks are shown in Table \ref{tab:unsup_claif}. 
We observe that CLAIF achieves the best performance on the four datasets STS15, STS16, STS-B, SICK-R and get the highest averaged Spearman's correlation on seven STS datasets.
And in the comparison with dataset generation methods, CLAIF outperforms Dino by 3.37 and 2.75 points on BERT and RoBERTa.
Therefore, we believe that CLAIF is more effective for the learning of sentence embeddings than CLZF methods.

We implement CLHAIF by incorporating AI feedback into supervised SimCSE and supervised PromptBERT/PromptRoBERTa.
We compare CLHAIF with other methods that use additional labeled datasets for training.
As shown in Table \ref{tab:sup_claif},
incorporating AI feedback improves results of CLHF methods like supervised SimCSE on six STS datasets except STS12.

\noindent \textbf{Transfer Tasks} \ In addition to STS tasks, we also evaluate several transfer learning tasks from SentEval.
Experimental results show that sentence embeddings learned with CLAIF and CLHAIF also achieve better or comparable performance compared to baselines.
We present the average results for seven transfer tasks in Table \ref{tab:result_transfer_avg} and detailed results in Appendix \ref{sec:transfer_tasks}.

\begin{table*}[!t]
\centering
\renewcommand\arraystretch{1.1}
\small
\setlength{\tabcolsep}{9.5pt}

\begin{tabular}{lcccccccc}
\hline
\textbf{Model} & \textbf{STS12} & \textbf{STS13} & \textbf{STS14} & \textbf{STS15} & \textbf{STS16} & \textbf{STS-B} & \textbf{SICK-R} & \textbf{Avg.}\\
\hline
\hline
\multicolumn{9}{c}{\textit{BERT-base}} \\
\hline
Trans-Encoder$_{\textrm{cross}}$ &
71.94 &
84.14 &
76.39 &
82.87 &
80.65 &
81.06 &
71.16 &
78.32 \\
CLAIF$_{\textrm{cross}}$ & 70.36 & 83.27 & 79.73 & 87.87 & 84.54 & 85.00 & 78.33 & 81.30 \\
% CLAIF$_{\textrm{bi}}$ & 70.62 & 81.51 & 76.29 & 85.05 & 81.36 & 84.34 & 78.22 & 79.63 \\
% CLAIF$_{\textrm{cross-to-bi}}$ & 71.41$_{\uparrow 0.79}$ & 82.77$_{\uparrow 1.26}$ & 77.07$_{\uparrow 0.78}$ & 85.79$_{\uparrow 0.74}$ & 82.46$_{\uparrow 1.10}$ & 84.75$_{\uparrow 0.41}$ & 77.84$_{\downarrow 0.38}$ & 80.30$_{\uparrow 0.67}$ \\
\hline
\hline
\multicolumn{9}{c}{\textit{RoBERTa-base}} \\
\hline
Trans-Encoder$_{\textrm{cross}}$ &
72.59 &
83.24 &
76.83 &
84.20 &
82.82 &
82.85 &
69.51 &
78.86 \\
CLAIF$_{\textrm{cross}}$ & 72.80 & 83.75 & 81.52 & 88.66 & 86.61 & 87.05 & 81.28 & 83.10 \\
% CLAIF$_{\textrm{bi}}$ & 68.33 & 82.26 & 77.00 & 85.18 & 83.43 & 85.05 & 78.02 & 79.90 \\
% CLAIF$_{\textrm{cross-to-bi}}$ & 71.75 $_{\uparrow 3.42}$ & 83.38$_{\uparrow 1.12}$ & 78.77$_{\uparrow 1.77}$ & 85.90$_{\uparrow 0.72}$ & 84.80$_{\uparrow 1.37}$ & 86.25$_{\uparrow 1.20}$ & 79.73$_{\uparrow 1.71}$ & 81.51$_{\uparrow 1.61}$ \\
\hline
\end{tabular}
\caption{The performance comparison of CLAIF based on the cross-encoder architecture.
} \label{tab:sts_cross_to_bi}
\end{table*}

\subsection{Scalability of CLAIF}
In this section we discuss the scalability of CLAIF.
The results of CLAIF$_{\textrm{scaled}}$ in Table \ref{tab:sup_claif} show that using more data to scale CLAIF can bring significant improvements.
CLAIF$_{\textrm{scaled}}$ greatly outputforms CLAIF by 2.74 points on BERT-base (79.63 $\rightarrow$ 82.37 ) and even outputforms or performs on par with CLHF and CLHAIF methods.
We believe that using more data can further improve the performance of CLAIF.
Since collecting data from AI feedback is more cheaper than from human feedback, we argue that CLAIF has great potential in practical applications.
% In Appendix \ref{sec:cost}, we present the total cost of using GPT-3 for our experiments.

\subsection{Sentence-Pair Modeling}
In this section, we evaluate CLAIF on the sentence-pair modeling task.
Cross-encoders usually outperform bi-encoders in information retrieval.
However, we observe in \citet{TransEncoder} that the cross-encoder does not show its superior on sentence-pair modeling.
We contribute this to the lack of fine-grained training signals.
We train a cross-encoder with CLAIF.
Experimental results in Table \ref{tab:sts_cross_to_bi} show that, with the help of AI feedback, CLAIF$_{\textrm{cross}}$ brings significant improvements for cross-encoders on the sentence-pair modeling task compared to the previous model Trans-Encoder \cite{TransEncoder}. 
More training details are in Appendix \ref{sec:cross_to_bi}.

% Experimental results in Table \ref{tab:sts_cross_to_bi} show that using the cross-encoder architecture achieves a substantial improvement compared to the bi-encoder.
% And adding the local AI feedback from the cross-encoder to CLAIF further improves the performance of the bi-encoder, which can generate better sentence embeddings.

\subsection{Human Evaluation}
In this section, we conduct human evaluation to measure the quality of generated sentences and similarity scores.
We measure whether the generated sentences are fluent and whether the similarity scores are consistent with the real semantic similarities.
% of their corresponding sentence pairs.
To help human judge the consistency, we generate a natural language explanation for each generated similarity score using GPT-3.
We invite 4 experts to participate in our human evaluation.
Then we random pick 100 samples from the dataset used in CLAIF and assign 25 samples to each expert.
In the evaluation, 92 percent of generated sentences are considered fluent and 90 percent of generated scores are considered consistent by the expert, which means our method can generate high quality sentence pairs and similarity scores.

\section{Related Work}

% \subsection{Sentence Embeddings}
% Recent studies about sentence embeddings can be divided into two lines of work.
% The first line of work focuses on using post-processing methods for pre-trained language models to mitigate the representation degeneration problem \cite{DBLP:conf/iclr/GaoHTQWL19}.
% \citet{BERT-Flow} use normalizing flows to transform the anisotropic sentence embedding distribution to a isotropic Gaussian distribution.
% \citet{BERT-Whitening} use the whitening operation to enhance the isotropy of sentence representations.
% \citet{PromptBERT} use prompts and take the \mask\ representation as the sentence embedding to avoid embedding biases.

Recent studies about sentence embeddings mainly focuse on using additional data to further train pre-trained language models.
% These methods usually achieve better performance than post-processing methods.
\citet{ConSERT} and \citet{SimCSE} propose different data augmentation strategies for contrastive learning and achieve significant improvements using unlabeled data.
\citet{DiffCSE} use equivariant contrastive learning for learning better representations.
\citet{Debiased_Negatives} and \citet{ESimCSE} address the bias caused by construction processes of negative and positive samples.
\citet{PromptBERT} use different prompt templates to produce positive pairs for contrastive learning.
\citet{SSSBERT} use various semantic sentence features to construct fine-grained labels for sentence embedding training.

% These methods usually achieve better performance when training with labeled datasets of other tasks like natural language inference, which is considered as the supervised setting.

% \subsection{Contrastive Learning}
% Contrastive learning is a self-supervised learning method for representation learning.

% \subsection{Dataset Generation}
Impressed by the powerful capabilities of LLMs \cite{GPT3,InstructGPT}, researchers pay more attention to using AI feedback from LLMs for zero-shot and few-shot learning.
\citet{UnifiedICL, MoT} use AI feedback from language models to enhance In-context Learning and Chain-of-Thoughts.
\citet{ZeroGen} and \citet{SuperGen} generate datasets by taking labels and prompts as the input of LLMs and then let LLMs generate training samples.
\citet{Dino} design a dataset generation method for STS tasks. 
They construct three natural language instructions based on three discrete similarity scores and then use these instructions to steer LLMs to construct sentence pairs.
However, it is hard to use natural language to describe various similarity scores, since the similarity score is a continuous variable with values ranging from 0 to 1.

\section{Conclusion}

In this paper, we first formalize four types of contrastive learning: contrastive learning from zero feedback (CLZF), contrastive learning from human feedback (CLHF), contrastive learning from AI feedback (CLAIF) and contrastive learning from human and AI feedback (CLHAIF).
Then we improve contrastive learning of sentence embeddings from AI feedback and combine human feedback with AI feedback to produce better supervision signals.
Experimental results show that CLAIF and CLHAIF can bring substantial improvements for sentence embedding learning.
We hope that learning from AI feedback can shed new lights for representation learning and contrastive learning.

\section*{Limitations}
To inspire future work, we conclude some limitations of our work as follows:
\begin{itemize}
    \item While our method achieves promising performance on sentence embedding related tasks like STS, the performance on other natural language processing tasks are still need to investigate.
    \item The AI feedback in our experiments comes from GPT-3, which requires a fee to use.
    \item We do not explore the effect of different task description prompts on the quality of generated sample pairs, which may influence the performance of CLAIF.
    \item In CLHAIF, we only use the AI feedback for positive sample pairs. How to utilize AI feedback for negative sample pairs remains to be studied.
\end{itemize}

\section*{Acknowledgement}
We would like to thank all anonymous reviewers for their valuable advice.
This work was supported by the National Natural Science Foundation of China (No. 62236004 and No. 62022027).

\bibliography{anthology,custom}
\bibliographystyle{acl_natbib}

\appendix

\section{Implementation Details}
\label{sec:training_details}

% Our code is based on sentence-transformers \cite{SentenceBERT} and Dino \cite{Dino}.

For CLAIF, we train our models for 3 epochs with a batch size of 32, and set the learning rate to 2e-5.
Following previous work, we use the development set of STS-B as the validation set.
We evaluate the model every 125 training steps on the validation set to choose the best checkpoint during training. 
We conduct a grid-search of learning rate $\in$ \{1e-5,2e-5\} on the validation set.

For CLHAIF, we use the official implementation and the default configuration of our baselines SimCSE \cite{SimCSE} and PrompBERT \cite{PromptBERT}. 
We only replace the one-hot label with our soft label.

We run experiments of CLAIF on a single RTX 3090 GPU with 24G gpu memory and experiments of CLHAIF on 4 RTX 3090 GPUs.
We fix the random seed to 42 for all experiments.

\section{Task Descriptions}
\label{sec:task_descriptions}
We use three task description prompts in our experiments.
For sentence pair generation in Section \ref{sec:sentence_pair_generation}, our two prompts are:

\noindent \emph{"Replace all <mask> tokens in '<masked-sentence>' to make a new sentence. The new sentence is:"} and \emph{"Write two sentences that mean the same thing. Sentence 1: '<sentence1>' Sentence 2:"}.

\noindent For semantic similarity labeling in Section \ref{sec:semantic_similarity_labeling}, our prompt is:

\noindent \emph{"The similarity score for two sentences is in the range from 0.0 to 1.0, 0.0 means completely different and 1.0 means almost the same. Now given two sentences '<sentence1>' and '<sentence2>', please give a similarity score for these two sentences: The similarity score for these two sentences is"}.

% \section{Cost for GPT-3}
% \label{sec:cost}

\section{Transfer Learning Tasks}
\label{sec:transfer_tasks}
We list the detailed performance comparison of CLAIF and CLHAIF in Table \ref{tab:transfer_tasks_unsupcl} and Table \ref{tab:transfer_tasks_supcl}.
Experimental results show that CLAIF achieves the best performance on RoBERTa-base and comparable performance on BERT-base. 
CLHAIF also achieves better results compared to the baselines.
Using more data to scale CLAIF also brings performance improvements on transfer learning tasks as shown in Tabel \ref{tab:transfer_tasks_supcl}.

\begin{table*}[t]
\renewcommand\arraystretch{1.2}
\centering
\small
\setlength{\tabcolsep}{8pt}
\begin{tabular}{lcccccccc}
\hline
\textbf{Model} & \textbf{MR} & \textbf{CR} & \textbf{SUBJ} & \textbf{MPQA} & \textbf{SST-2} & \textbf{TREC} & \textbf{MRPC} & \textbf{Avg.}\\
\hline
\hline
\multicolumn{9}{c}{\textit{BERT-base}} \\
\hline
Avg. BERT embeddings\(^\dagger\) & 78.66 & 86.25 & 94.37 & 88.66 & 84.40 & \textbf{92.80} & 69.54 & 84.94 \\
BERT- [CLS] embedding\(^\dagger\) & 78.68 & 84.85 & 94.21 & 88.23 & 84.13 & 91.40 & 71.13 & 84.66 \\
SimCSE\(^\ddagger\) & 81.18 & 86.46 & 94.45 & 88.88 & 85.50 & 89.80 & 74.43 & 85.81 \\
SimCSE w/MLM\(^\ddagger\) & \textbf{82.92} & 87.23 & \textbf{95.71} & 88.73 & \textbf{86.81} & 87.01 & \textbf{78.07} & 86.64 \\
DiffCSE\(^\ddagger\) & 82.69 & 87.23 & 95.23 & 89.28 & 86.60 & 90.40 & 76.58 & \textbf{86.86} \\
PromptBERT\(^\dagger\) & 80.74 & 85.49 & 93.65 & 89.32 & 84.95 & 88.20 & 76.06 & 85.49\\
\hline
Dino$_{\textrm{GPT-3}}$ & 79.96 & 85.27 & 93.67 & 88.87 & 84.29 & 88.60 & 69.62 & 84.33 \\
\hline
CLAIF & 81.64 & \textbf{87.98} & 94.24 & \textbf{89.34} & 86.16 & 89.80 & 77.16 & 86.62 \\
\hline
\hline
\multicolumn{9}{c}{\textit{RoBERTa-base}} \\
\hline
Avg. RoBERTa embeddings & \textbf{84.35} & 88.34 & \textbf{95.28} & 86.13 & 89.46 & \textbf{93.20} & 74.20 & 87.28 \\
SimCSE\(^\ddagger\) & 81.04 & 87.74 & 93.28 & 86.94 & 86.60 & 84.60 & 73.68 & 84.84 \\
SimCSE w/MLM\(^\ddagger\) & 83.37 & 87.76 & 95.05 & 87.16 & 89.02 & 90.80 & 75.13 & 86.90 \\
DiffCSE\(^\ddagger\) & 82.82 & 88.61 & 94.32 & 87.71 & 88.63 & 90.40 & 76.81 & 87.04 \\
PromptRoBERTa\(^\dagger\)& 83.82 & 88.72 & 93.19 & \textbf{90.36} & 88.08 & 90.60 & 76.75 & 87.36 \\
\hline
Dino$_{\textrm{GPT-3}}$ & 82.31 & 88.66 & 93.95 & 88.72 & 87.53 & 88.20 & 73.74 & 86.16 \\
\hline
% Ours & \textbf{84.89} & \textbf{90.76} & 94.52 & 88.86 & \textbf{90.33} & \textbf{92.20} & \textbf{77.51} & \textbf{88.44} \\
CLAIF & 84.11 & \textbf{90.62} & 94.29 & 89.13 & \textbf{89.57} & 91.00 & \textbf{77.22} & \textbf{87.99} \\
\hline
\end{tabular}
\caption{The performance comparison of CLAIF on transfer learning tasks.
\(\dagger\): results from \cite{PromptBERT}. \(\ddagger \): results from \cite{DiffCSE}. Other results are from our experiments.
} \label{tab:transfer_tasks_unsupcl}
\end{table*}

\begin{table*}[t]
\renewcommand\arraystretch{1.2}
\centering
\small
\setlength{\tabcolsep}{4pt}
\begin{tabular}{lcccccccc}
\hline
\textbf{Model} & \textbf{MR} & \textbf{CR} & \textbf{SUBJ} & \textbf{MPQA} & \textbf{SST-2} & \textbf{TREC} & \textbf{MRPC} & \textbf{Avg.}\\
\hline
\hline
\multicolumn{9}{c}{\textit{BERT-base}} \\
\hline
SBERT\(^\dagger\) & \textbf{83.64} & \textbf{89.43} & 94.39 & 89.86 & \textbf{88.96} & 89.60 & 76.00 & \textbf{87.41} \\
SimCSE\(^\dagger\) & 82.69 & 89.25 & \textbf{94.81} & 89.59 & 87.31 & 88.40 & 73.51 & 86.51 \\
\hspace{1em} w/ CLHAIF & 
83.11$_{\uparrow 0.42}$ & 
88.98$_{\downarrow 0.27}$ & 
94.47$_{\downarrow 0.34}$ & 
89.95$_{\uparrow 0.36}$ & 
88.58$_{\uparrow 1.27}$ & 
86.40$_{\downarrow 2.00}$ & 
75.65$_{\uparrow 2.14}$ & 
86.73$_{\uparrow 0.22}$ \\
PromptBERT$^*$ & 83.05 & 88.96 & 94.68 & 89.86 & 88.19 & 87.80 & 76.29 & 86.98 \\
\hspace{1em} w/ CLHAIF & 
83.14$_{\uparrow 0.09}$ & 
89.12$_{\uparrow 0.16}$ & 
94.65$_{\downarrow 0.03}$ & 
89.97$_{\uparrow 0.11}$ & 
87.86$_{\downarrow 0.33}$ & 
88.80$_{\uparrow 1.00}$ & 
76.06$_{\downarrow 0.23}$ & 
87.09$_{\uparrow 0.11}$ \\
\hline
CLAIF$_{\textrm{scaled}}$ & 82.08 & 89.12 & 94.48 & \textbf{90.22} & 87.53 & \textbf{90.20} & \textbf{76.41} & 87.15 \\
\hline
\hline
\multicolumn{9}{c}{\textit{RoBERTa-base}} \\
\hline
SRoBERTa\(^\dagger\) & 84.91 & 90.83 & 92.56 & 88.75 & 90.50 & 88.60 & 78.14 & 87.76 \\
SimCSE\(^\dagger\) & 84.92 & \textbf{92.00} & 94.11 & 89.82 & 91.27 & 88.80 & 75.65 & 88.08 \\
\hspace{1em} w/ CLHAIF & 
86.10$_{\uparrow 1.18}$ & 
91.76$_{\downarrow 0.24}$ & 
94.66$_{\uparrow 0.55}$ & 
90.07$_{\uparrow 0.25}$ & 
91.93$_{\uparrow 0.66}$ & 
91.60$_{\uparrow 2.80}$ & 
75.59$_{\downarrow 0.06}$ & 
88.82$_{\uparrow 0.74}$ \\
PromptRoBERTa$^*$ & 86.22 & 91.55 & \textbf{95.08} & 90.97 & 91.82 & 91.40 & 76.70 & 89.11 \\
\hspace{1em} w/ CLHAIF & 
\textbf{86.41}$_{\uparrow 0.19}$ & 
91.76$_{\uparrow 0.21}$ & 
94.90$_{\downarrow 0.18}$ & 
\textbf{91.01}$_{\uparrow 0.04}$ & 
\textbf{92.04}$_{\uparrow 0.22}$ & 
92.40$_{\uparrow 1.00}$ & 
76.35$_{\downarrow 0.35}$ & 
89.27$_{\uparrow 0.16}$ \\
\hline
CLAIF$_{\textrm{scaled}}$ & 85.05 & 91.71 & 94.39 & 90.03 & 91.87 & \textbf{94.00} & \textbf{79.01} & \textbf{89.44} \\
\hline
\end{tabular}
\caption{The performance comparison of CLHAIF on transfer learning tasks.
\(\dagger\): results from \citet{PromptBERT}.
$*$: The results of PromptBERT and PromptRoBERTa are obtained by running official code of \citet{PromptBERT} with recommended hyperparameters.
} \label{tab:transfer_tasks_supcl}
\end{table*}

\section{Sentence-Pair Modeling}
\label{sec:cross_to_bi}
In sentence-pair modeling task, cross-encoders can be used to directly encode the sequence of two sentences and then predict a similarity score.
Previous studies \cite{Augmented_SBERT, TransEncoder, ERNIE-Search} show that cross-encoders usually outperform bi-encoders.
We find that CLAIF can significantly improve the performance of cross-encoders on sentence-pair modeling task, with the help of fine-grained AI feedback scores.

We use the binary cross-entropy (BCE) loss to train cross-encoders initialized from BERT and RoBERTa:
\begin{gather}
    \mathcal{L}=-\frac{1}{N}\sum_{i=1}^{N}l_i \\
    l_i=y_{i}\log{\sigma(\hat{y_i})}+(1-y_{i})\log{(1-\sigma(\hat{y_i}))} \nonumber
\end{gather}
where $N$ is the batch size,
$\hat{y_i}$ is the predicted score of the i-th sentence pair,
$y_i$ is the AI feedback similarity score and $\sigma$ is the sigmoid function.

\section{Cost for Data Generation}
According to our billings, we spent about \$100 to generate data for CLAIF and about \$720 for the scaled dataset.

\section{Generated Examples}
\label{sec:generated_examples}
We present some generated sample pairs used in CLAIF in Table \ref{tab:claif_exp} and some generated similarity scores for sample pairs constructed from NLI in Table \ref{tab:clhaif_exp}.

\begin{table*}[h]
\centering
% \fontsize{8}{9}\selectfont
\begin{tabular}{llc}
\toprule
\textbf{Original Sentence} & \textbf{Generated Sentence} & \textbf{Similarity Score} \\
\hline
\multirow{1}{*}{\begin{tabular}[c]{@{}l@{}} a plane is taking off . \\ \end{tabular}} &

\begin{tabular}[c]{@{}l@{}}
an aircraft is departing . \\  
The airplane is taking off. \\
A plane is taking off swiftly \\
The blue plane is taking off. \\
Airplane is flying. \\ 
Bob and Joe are taking a walk. \\
Aeroplane is flying \\
Put off steam \\
Turn off lights \\
\end{tabular} 
& 
\begin{tabular}[c]{@{}c@{}} 
0.80 \\
0.80 \\
0.90 \\
0.75 \\
0.67 \\
0.00 \\
0.67 \\
0.00 \\
0.00 \\
\end{tabular} \\ 
\hline
\multirow{1}{*}{\begin{tabular}[c]{@{}l@{}} a man is playing a large flute . \\ \end{tabular}} &

\begin{tabular}[c]{@{}l@{}}
A male individual is performing on a big flute. \\  
a man is playing a large flute. \\
He she is playing a large flute. \\
a man played a wooden flute. \\
a flute is not a wooden flute \\ 
a boy playing a large drum \\
a man is wise. \\
The old man stood . \\
The quick brown fox jumps over the lazy dog \\
\end{tabular} 
& 
\begin{tabular}[c]{@{}c@{}} 
0.86 \\
1.00 \\
0.78 \\
0.71 \\
0.20 \\
0.33 \\
0.00 \\
0.00 \\
0.00 \\
\end{tabular} \\ 
\hline
\multirow{1}{*}{\begin{tabular}[c]{@{}l@{}} three men are playing chess . \\ \end{tabular}} &

\begin{tabular}[c]{@{}l@{}}
There are three men playing chess. \\  
Three children are playing chess. \\
Three kings are playing chess. \\
They are playing chess . \\
three men played chess together \\ 
three men are walking \\
John and Mary were playing chess together \\
I play blitz chess online \\
I like to play soccer and tennis. \\
\end{tabular} 
& 
\begin{tabular}[c]{@{}c@{}} 
0.94 \\
0.80 \\
0.87 \\
0.80 \\
0.78 \\
0.00 \\
0.50 \\
0.20 \\
0.00 \\
\end{tabular} \\ 
\bottomrule
\end{tabular}
\caption{Generated examples of sample pairs used in CLAIF.}
\label{tab:claif_exp}
\end{table*}

\begin{table*}[h]
\centering
% \fontsize{8}{9}\selectfont
\begin{tabular}{llc}
\toprule
\textbf{Premise} & \textbf{Entailment Hypothesis} & \textbf{Similarity Score} \\
\hline
The other men shuffled. & 
The other men were shuffled around. &
0.78 \\
\hline
well it's been very interesting &
It has been very intriguing. &
0.90 \\
\hline
He started slowly back to the bunkhouse. &
He returned slowly to the bunkhouse. &
0.91 \\
\hline
well what the market can bear and &
The market can bear some. &
0.71 \\
\hline
She smiled back. &
She was happy. &
0.25 \\
\hline
The economy could be still better. &
It still have room for improvement. &
0.55 \\
\hline
The man should have died instantly. &
The man should not have been alive. & 
0.14 \\
\hline
Turned out, I wasn't completely wrong. &
I was not totally wrong. &
0.8 \\
\bottomrule
\end{tabular}
\caption{Generated examples of similarity scores used in CLHAIF.}
\label{tab:clhaif_exp}
\end{table*}

\section{Comparison with Text-Ada-Embedding-002}
\label{sec:ada}
Recently, OpenAI has released a powerful embedding model named text-ada-embedding-002, we compare the performance of it on STS tasks with CLAIF here.
The results show that CLAIF-scaled achieves better performance on STS tasks than text-ada-embedding-002.

\begin{table*}[!t]
\centering
\renewcommand\arraystretch{1.1}
\small
\setlength{\tabcolsep}{9.5pt}

\begin{tabular}{lcccccccc}
\toprule
\textbf{Model} & \textbf{STS12} & \textbf{STS13} & \textbf{STS14} & \textbf{STS15} & \textbf{STS16} & \textbf{STS-B} & \textbf{SICK-R} & \textbf{Avg.}\\
\midrule
Ada-Embedding-002 &
69.80 &
83.26 &
76.08 &
86.12 &
\textbf{85.96} &
84.30 &
\textbf{80.25} &
80.82 \\
CLAIF-BERT & 70.62 & 81.51 & 76.29 & 85.05 & 81.36 & 84.34 & 78.22 & 79.63 \\
CLAIF-BERT$_{\textrm{scaled}}$ &
\textbf{74.36} &
\textbf{85.07} &
\textbf{80.64} &
\textbf{87.21} & 
83.36 & 
\textbf{86.26} &
79.68 & 
\textbf{82.37} \\
\bottomrule
\end{tabular}
\caption{The performance comparison between CLAIF and OpenAI's text-ada-embedding-002.
} \label{tab:sts_cross_to_bi}
\end{table*}

\end{document}